\documentclass{jmlr}

\jmlrvolume{1}
\jmlryear{2018}
\jmlrworkshop{ICML 2018 AutoML Workshop}

\usepackage{subcaption}
\usepackage{dsfont}
\usepackage{arydshln}

\title[Backprop Evolution]{Backprop Evolution}

 \author{\Name{Maximilian Alber} \nametag{\thanks{Contributed equally.} \thanks{Work was done as intern at Google Brain.}} \Email{maximilian.alber@tu-berlin.de}\\
  \addr TU Berlin
  \AND
  \Name{Irwan Bello} \nametag{\footnotemark[1]} \Email{ibello@google.com}\\
  \Name{Barret Zoph} \Email{barretzoph@google.com}\\
  \Name{Pieter-Jan Kindermans} \nametag{\thanks{Work was done as a member of the Google AI Residency program (\href{http://g.co/airesidency}{g.co/airesidency}).}} \Email{pikinder@google.com}\\
  \Name{Prajit Ramachandran} \nametag{\footnotemark[3]} \Email{prajit@google.com}\\
  \Name{Quoc Le} \Email{qvl@google.com}\\
  \addr Google Brain
 }

\newcommand{\towrite}[1]{ }

\newcommand{\norm}[1]{ #1/\Vert.\Vert}
\newcommand{\enorm}[1]{#1/\Vert.\Vert^{elem}}
\newcommand{\colnorm}[1]{#1/\Vert.\Vert^{col}}
\newcommand{\rownorm}[1]{#1/\Vert.\Vert^{row}}
\newcommand{\rownormx}[1]{#1/\Vert.\Vert^{row}}

\DeclareMathOperator{\gnoise}{\text{gnoise}}
\DeclareMathOperator{\sgn}{\text{sgn}}
\DeclareMathOperator{\clip}{\text{clip}}
\DeclareMathOperator{\drop}{\text{drop}}
\DeclareMathOperator{\deltai}{b^p_{i+1} \frac{\partial h^p_{i+1}}{\partial  h^p_{i}}}
\DeclareMathOperator{\errhi}{b^p_{i+1} \frac{\partial h^p_{i+1}}{\partial  h_{i}}}
\DeclareMathOperator{\deltalastb}{b^p_{L} \cdot R_{L i}}
\DeclareMathOperator{\directfeedback}{(b^p_{L} \cdot R_{L i}) \odot \frac{\partial h_{i}}{\partial  h^p_{i}}}

\DeclareMathOperator{\gprimei}{\frac{\partial h_i}{\partial  h^p_{i}}}
\DeclareMathOperator{\rmean}{\hat{\text{m}}}
\DeclareMathOperator{\rstd}{\hat{\text{s}}}
\DeclareMathOperator{\rbn}{\text{bn}}

\begin{document}

\maketitle

\begin{abstract}
  The back-propagation algorithm is the cornerstone of deep learning. Despite its importance, few variations of the algorithm have been attempted. This work presents an approach to discover new variations of the back-propagation equation. 
  We use a domain specific language to describe update equations as a list of primitive functions.
  An evolution-based method is used to discover new propagation rules that maximize the generalization performance after a few epochs of training.
  We find several update equations that can train faster with short training times than standard back-propagation, and perform similar as standard back-propagation at convergence.
\end{abstract}


\section{Introduction}

The back-propagation algorithm is one of the most important algorithms in machine learning (\cite{lin70,wer74,rumelhart1986learning}). A few attempts have been made to change the back-propagation equation with some degrees of success (e.g.,~\cite{bengio1994use,lillicrap2014random,lee2015difference,nokland2016direct,liao2016important}). Despite these attempts, modifications of back-propagation equations have not been widely used as these algorithms rarely improve practical applications, and sometimes hurt them.

Inspired by the recent successes of automated search methods for machine learning~\citep{zoph2016neural,zoph2017learning,bello2017neural,brock2017smash,real2018regularized,pmlr-v80-bender18a}, we propose a method for automatically generating back-propagation equations. To that end, we introduce a domain specific language to describe such mathematical formulas as a list of primitive functions and use an evolution-based method to discover new propagation rules. The search is conditioned to maximize the generalization after a few epochs of training. We find several variations of the equation that can work as well as the standard back-propagation equation. Furthermore, several variations can achieve improved accuracies with short training times. This can be used to improve algorithms like Hyperband \citep{li2017hyperband} which make accuracy-based decisions over the course of training.

\section{Backward propagation} 
\begin{figure}[h]
\centering
\includegraphics[width=\textwidth]{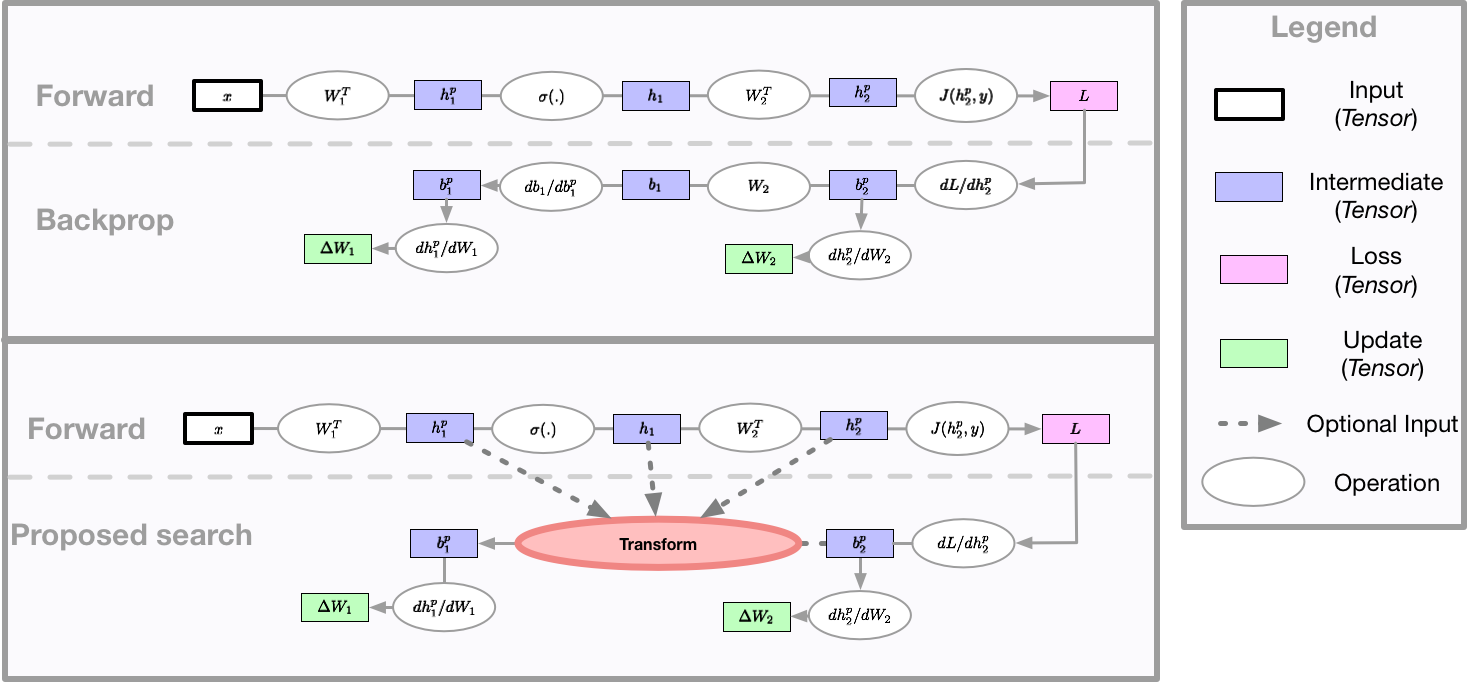}
\caption{Neural networks can be seen as computational graphs. The forward graph is defined by the network designer, while the back-propagation algorithm implicitly defines a computational graph for the parameter updates. Our main contribution is exploring the use of evolution to find a computational graph for the parameter updates that is more effective than standard back-propagation.}
\label{fig}
\end{figure}
%
The simplest neural network can be defined as a sequence of matrix multiplications and nonlinearities:
\begin{align*}
    h^p_{i} = W_i^T h_{i-1}, &\qquad\qquad h_i = \sigma(h^p_i).
\end{align*}%
where $h_0 = x$ is the input to the network, $i$ indexes the layer and $W_i$ is the weight matrix of the i-th layer.
To optimize the neural network, we compute the partial derivatives of the loss $J(f(x), y)$ w.r.t. the weight matrices
$
\frac{\partial  J(f(x), y)}{\partial W_i}.
$
This quantity can be computed by making use of the chain rule in the back-propagation algorithm. 
To compute the partial derivative with respect to the hidden activations $b^p_i=\frac{\partial  J(f(x), y)}{h^p_i}$, a sequence of operations is applied to the derivatives:%
\begin{alignat}{2}
    b_{L} = \frac{\partial J(f(x), y)}{\partial h_{L}}, 
    &\qquad\qquad b^p_{i+1} = \;b_{i+1} \;\frac{\partial h_{i+1}}{\partial  h^p_{i+1}}, 
    &&\qquad\qquad\;\;\;\;b_{i} = \;b^p_{i+1} \;\frac{\partial h^p_{i+1}}{\partial  h_{i}}.
    \label{eq:backprop}
\end{alignat}%
Once $b^p_i$ is computed, the weights update can be computed as: $\Delta W_i = b^p_i \frac{\partial h^p_i}{W_{i}}$.

As shown in Figure \ref{fig}, the neural network can be represented as a forward and backward computational graph. Given a forward computational graph defined by a network designer, the back-propagation algorithm defines a backward computational graph that is used to update the parameters. However, it may be possible to find an improved backward computational graph that results in better \emph{generalization}.

Recently, automated search methods for machine learning have achieved strong results on a variety of tasks  (e.g.,~\cite{zoph2016neural,baker2016designing,real2017large,bello2017neural,brock2017smash,ramachandran2018searching,zoph2017learning,bender2018demystifying,real2018regularized}). With the exception of ~\cite{bello2017neural}, these methods involve modifying the forward computational graph, relying on back-propagation to define the appropriate backward graph. In this work, we instead concern ourselves with modifying the backward computational graph and use a search method to find better formulas for $b^p_i$, yielding new training rules.

\section{Method}

In order to discover improved update rules, we employ an evolution algorithm to search over the space of possible update equations. At each iteration, an evolution controller sends a batch of mutated update equations to a pool of workers for evaluation. Each worker trains a fixed neural network architecture using its received mutated equation and reports the obtained validation accuracy back to our controller.

\subsection{Search Space}
\label{DSL}

We use a domain-specific language (DSL) inspired by \citet{bello2017neural} to describe the equations used to compute $b^p_i$. 
The DSL expresses each $b_i^p$ equation as 
$f(u_1(op_1), u_2(op_2))$
where $op_1$, $op_2$ are possible operands, $u_1(\cdot)$ and $u_2(\cdot)$ are unary functions, and $f(\cdot,\cdot)$ is a binary function. The sets of unary functions and binary functions are manually specified but individual choices of functions and operands are selected by the controller. Examples of each component are as follows:
\begin{itemize}
    \item \textbf{Operands:}
    $W_i$ (weight matrix of the current layer),
    $R_i$ (Gaussian matrix),
    $R_{L i}$ (Gaussian random matrix mapping from $b^p_L$ to $b^p_i$),
    $h^p_i$, $h_i$, $h^p_{i+1}$ (hidden activations of the forward propagation),
    $b^p_L$, $b^p_{i+1}$ (backward propagated values).
    \item \textbf{Unary functions $u(x)$:}
    $x$ (identity), $x^t$ (transpose), $1/x$,
    $x^2$, $\sgn(x) x^2$, $x^3$, $a x$, $x + b$,
    $\drop_d(x)$ (dropout with drop probability $d \in (0.01, 0.1, 0.3)$),
    $\clip_c(x)$ (clip values in range $[-c, c]$ and $c \in (0.01, 0.1, 0.5, 1.0$)), 
    $\norm{x}_{fro}$, $\norm{x}_1$, $\norm{x}_{-inf}$, $\norm{x}_{inf}$ (normalizing term by matrix norm).
    \item \textbf{Binary functions $f(x, y)$:} $x + y$, $x - y$, $x \odot y$, $x / y$ (element-wise addition, subtraction, multiplication, division), $x \cdot y$ (matrix multiplication) and $x$ (keep left), $\min(x, y)$, $\max(x, y)$ (minimum and maximum of $x$ and $y$).
\end{itemize}
where $i$ indexes the current layer. The full set of components used in our experiments is specified in Appendix~\ref{appsearchspace}. The resulting quantity $f(u_1(op_1), u_2(op_2))$ is then either used as $b_i^p$ in Equation~\ref{eq:backprop} or used recursively as $op_1$ in subsequent parts of the equation. In our experiments, we explore equations composed of between 1 and 3 binary operations.
This DSL is simple but can represent complex equations such as normal back-propagation, feedback alignment \citep{lillicrap2014random}, and direct feedback alignment \citep{nokland2016direct}.

\subsection{Evolution algorithm} 

The evolutionary controller maintains a population of discovered equations. At each iteration, the controller does one of the following:
1) With probability $p$, the controller chooses an equation randomly at uniform within the $N$ most competitive equations found so far during the search,
2) With probability $1-p$, it chooses an equation randomly at uniform from the rest of the population.
The controller subsequently applies $k$ mutations to the selected equation, where $k$ is drawn from a categorical distribution. Each of these $k$ mutations simply consists in selecting one of the equation components (e.g., an operand, an unary, or a binary function) uniformly at random and swapping it with another component of the same category chosen uniformly at random. Certain mutations lead to mathematically infeasible equations (e.g., a shape mismatch when applying a binary operation to two matrices). When this is the case, the controller restarts the mutation process until successful. $N$, $p$ and the categorical distribution from which $k$ is drawn are hyperparameters of the algorithm.


To create an initial population, we simply sample $N$ equations at random from the search space. Additionally, in some of our experiments, we instead start with a small population of pre-defined equations (typically the normal back-propagation equation or its feedback alignment variants). The ability to start from existing equations is an advantage of evolution over reinforcement learning based approaches~(\cite{zoph2016neural,zoph2017learning,bello2017neural,ramachandran2018searching}).

\section{Experiments}

The choice of model used to evaluate each proposed update equation is an important setting in our method. Larger, deeper networks are more realistic but take longer to train, whereas small models are faster to train but may lead to discovering update equations that do not generalize. We strike a balance between the two criteria by using Wide ResNets (WRN) \citep{zagoruyko2016wide} with 16 layers and a width multiplier of 2 (WRN 16-2) to train on the CIFAR-10 dataset.

We experiment with different search setups, such as varying the subsets of the available operands and operations and using different optimizers.
The evolution controller searches for the update equations that obtain the best accuracy on the validation set.
We then collect the $100$ best discovered equations, which were determined by the mean accuracy of 5 reruns.
Finally, the best equations are used to train models on the full training set, and we report the test accuracy. The experiment setup is further detailed in Appendix~\ref{appexperimentaldetails}.

\begin{table}[h]
\hspace{-0.5cm}
\begin{tabular}{ | r | c | r | c | }
  \hline
  \multicolumn{2}{| c |}{\textbf{SGD}} & \multicolumn{2}{c|}{\textbf{SGD and Momentum}} \\
  \hline
  \multicolumn{4}{c}{}\\
  \multicolumn{4}{c}{\textbf{(A1) Search validation}}\\
  \hline
  \emph{baseline} $g^p_i$ & $77.11\pm3.53$ & \emph{baseline} $g^p_i$ & $83.00\pm0.63$ \\
  \hline
  {\tiny $\min(\norm{g^p_i}_{fro}), \clip_{1.0}(h_i))$ } & $\mathbf{84.48\pm0.45}$ &
  {\tiny $(\enorm{g^p_i}_2) / (2 + \rmean(\directfeedback))$ } & $\mathbf{85.43\pm1.59}$ \\
  {\tiny $(\rownormx{g^p_i}_2) + \enorm{\rmean(\gprimei)}_0$ } & $\mathbf{84.41\pm1.37}$ &
  {\tiny $(\norm{g^p_i}_{fro}) + 0.5 g^p_i$ } & $\mathbf{85.36\pm1.41}$ \\
  \hdashline
  {\tiny $\norm{g^p_i}_fro$ } & $\mathbf{\mathbf{84.15\pm0.66}}$ &
  {\tiny $\norm{g^p_i}_fro$ } & $\mathbf{84.23\pm0.88}$ \\
  {\tiny $\enorm{g^p_i}_2$ } & $\mathbf{83.16\pm0.90}$ &
  {\tiny $\enorm{g^p_i}_2$ } & $\mathbf{83.83\pm1.27}$ \\
  \hline
  \multicolumn{4}{c}{\textbf{(A2) Generalize to WRN 28x10}}\\
  \hline
  \emph{baseline} $g^p_i$ & $73.10\pm1.41$ & \emph{baseline} $g^p_i$ & $79.53\pm2.89$ \\
  \hline
  {\tiny $(\enorm{g^p_i}_{inf}) \odot (\norm{\rstd(\gprimei)}_1)$ } & $\mathbf{88.22\pm0.55}$ &
  {\tiny $\max((\errhi-10.0), \enorm{g^p_i}_2)$ } & $\mathbf{89.43\pm0.99}$ \\
  {\tiny $\clip_{0.01}(0.01 + h_i^p - (h_i^p)^+) \odot (\enorm{g^p_i}_{inf})$ } & $\mathbf{87.28\pm0.29}$ &
  {\tiny $(\errhi - 0.01) + (\enorm{g^p_i}_2)$ } & $\mathbf{89.26\pm0.67}$ \\
  \hdashline
  {\tiny $\norm{g^p_i}_{fro}$ } & $\mathbf{87.17\pm0.87}$ &
  {\tiny $\norm{g^p_i}_fro$ } & $\mathbf{89.63\pm0.32}$ \\
  {\tiny $\enorm{g^p_i}_2$ } & $\mathbf{85.30\pm1.04}$ &
  {\tiny $\enorm{g^p_i}_2$ } & $\mathbf{89.05\pm0.88}$ \\
  \hline
  \multicolumn{4}{c}{\textbf{(A3) Generalize to longer training}}\\
  \hline
  \emph{baseline} $g^p_i$ & $92.38\pm0.10$ & \emph{baseline} $g^p_i$ & $93.75\pm0.15$ \\
  \hline
  {\tiny $(\enorm{g^p_i}_2) \odot \sgn(\rbn(\gprimei))$ } & $\mathbf{92.97\pm0.18}$ &
  {\tiny $\drop_{0.01}(g^p_i) - (\enorm{\rbn(\deltalastb)}_0)$ } & $93.72\pm0.20$ \\
  {\tiny $(\enorm{g^p_i}_2) - (\norm{\rmean(\errhi)}_{inf})$ } & $\mathbf{92.90\pm0.13}$ &
  {\tiny $(1 + \gnoise_{0.01}) g^p_i + (\enorm{g^p_ip1_b}_1)$ } & $93.66\pm0.12$ \\
  \hdashline
  {\tiny $\norm{g^p_i}_{fro}$ } & $\mathbf{92.85\pm0.14}$ &
  {\tiny $\norm{g^p_i}_{fro}$ } & $93.41\pm0.18$ \\
  {\tiny $\enorm{g^p_i}_2$ } & $\mathbf{92.78\pm0.13}$ &
  {\tiny $\enorm{g^p_i}_2$ } & $93.35\pm0.15$ \\
  \hline
  \multicolumn{4}{c}{\textbf{(B1) Searching with longer training}}\\
  \hline
  \emph{baseline} $g^p_i$ & $87.13\pm0.25$ & \emph{baseline} $g^p_i$ & $88.94\pm0.11$ \\
  \hline
  {\tiny $(\norm{g^p_i}_1) + (\enorm{\errhi}_2)$ } & $\mathbf{87.94\pm0.22}$ &
  {\tiny $2 g^p_i + (\enorm{\rbn(\errhi)}_1)$ } & $89.04\pm0.25$ \\
  {\tiny $(\norm{g^p_i}_1) \odot \clip_{0.1}(\gprimei)$ } & $\mathbf{87.88\pm0.39}$ &
  {\tiny $(\gprimei - 0.5) \odot 2.0 g^p_i$ } & $88.95\pm0.16$ \\
  {\tiny $(\enorm{g^p_i}_2) \odot \sgn(\rbn(\gprimei))$ } & $\mathbf{87.82\pm0.19}$ &
  {\tiny $(\enorm{\rstd(\errhi)}_{inf}) * 2.0 g^p_i$ } & $88.94\pm0.20$ \\
  {\tiny $(0.5 g^p_i) / (\rstd(\gprimei) + 0.1)$ } & $\mathbf{87.72\pm0.25}$ &
  {\tiny $(\gprimei)^+ \odot \clip_{1.0}(\errhi)$ } & $88.93\pm0.14$ \\
  \hline

\end{tabular}
\caption{Results of the experiments. For A1-3 we show the two best performing equations on each setup and two equations that consistently perform well across all setups. For B1 we show the four best performing equations. All results are the average test accuracy over $5$ repetitions. Baseline is gradient back-propagation. Numbers that are at least $0.1\%$ better are in bold. A description of the operands and operations can be found in  Appendix~\ref{appsearchspace}. We denote $\deltai$ with $g^p_i$.}

\label{table}
\end{table}


\subsection{Baseline search and generalization}

In the first search we conduct, the controller proposes update equations to train WRN 16-2 networks for 20 epochs with SGD with or without momentum. The top 100 update equations according to validation accuracy are collected and then tested on different scenarios:
\begin{description}
\item [(A1)] WRN 16-2 for 20 epochs, replicating the search settings.
\item [(A2)] WRN 28-10 for 20 epochs, testing generalization to larger models (WRN 28-10 has 10 times more parameters than WRN 16-2).
\item [(A3)] WRN 16-2 for 100 epochs, testing generalization to longer training regimes.
\end{description}
The results are listed in Table~\ref{table}. When evaluated on the search setting (A1), it is clear that the search finds update equations that outperform back-propagation for both SGD and SGD with momentum, demonstrating that the evolutionary controller can successfully find update equations that achieve better accuracies. The update equations also successfully generalize to the larger WRN 28-10 model (A2), outperforming the baseline by up to 15\% for SGD and 10\% for SGD with momentum. This result suggests that the usage of smaller models during searches is appropriate because the update equations still generalize to larger models.

However, when the model is trained for longer (A3), standard back-propagation and the discovered update equations perform similarly. The discovered update equations can then be seen as equations that speed up training initially but do not affect final model performance, which can be practical in settings where decisions are made during the early stages of training to continue the experiment, such as some hyperparameter search schemes \citep{li2017hyperband}.




\subsection{Searching for longer training times}

The previous search experiment finds update equations that work well at the beginning of training but do not outperform back-propagation at convergence. The latter result is potentially due to the mismatch between the search and the testing regimes, since the search used 20 epochs to train child models whereas the test regime uses 100 epochs.

A natural followup is to match the two regimes. In the second search experiment, we train each child model for 100 epochs. To compensate for the increase in experiment time due to training for more epochs, a smaller network (WRN 10-1) is used as the child model. The use of smaller models is acceptable since update equations tend to generalize to larger, more realistic models (see (A2)).

The results are shown in Table~\ref{table} section (B1) and are similar to (A3), i.e., we are able to find update rules that perform moderately better for SGD, but the results for SGD with momentum are comparable to the baseline. The similarity between results of (A3) and (B1) suggest that the training time discrepancy may not be the main source of error. Furthermore, SGD with momentum is fairly unchanging to different update equations. Future work can analyze why adding momentum increases robustness.


\section{Conclusion and future work}

In this work, we presented a method to automatically find equations that can replace standard back-propagation. We use an evolutionary controller that operates in a space of equation components and tries to maximize the generalization of trained networks.
The results of our exploratory study show that for specific scenarios, there are equations that yield better generalization performance than this baseline, but more work is required to find an equation that performs better in general scenarios.
It is left to future work to distill patterns from equations found by the search, and research under which conditions and why they yield better performance.



\paragraph{Acknowledgements:} We would like to thank Gabriel Bender for his technical advice throughout this work and Simon Kornblith for his valuable feedback on the manuscript.

\bibliography{ref}

\begin{thebibliography}{23}
\providecommand{\natexlab}[1]{#1}
\providecommand{\url}[1]{\texttt{#1}}
\expandafter\ifx\csname urlstyle\endcsname\relax
  \providecommand{\doi}[1]{doi: #1}\else
  \providecommand{\doi}{doi: \begingroup \urlstyle{rm}\Url}\fi

\bibitem[Abadi et~al.(2016)Abadi, Barham, Chen, Chen, Davis, Dean, Devin,
  Ghemawat, Irving, Isard, et~al.]{abadi2016tensorflow}
Mart{\'\i}n Abadi, Paul Barham, Jianmin Chen, Zhifeng Chen, Andy Davis, Jeffrey
  Dean, Matthieu Devin, Sanjay Ghemawat, Geoffrey Irving, Michael Isard, et~al.
\newblock Tensorflow: A system for large-scale machine learning.
\newblock In \emph{OSDI}, volume~16, pages 265--283, 2016.

\bibitem[Baker et~al.(2017)Baker, Gupta, Naik, and Raskar]{baker2016designing}
Bowen Baker, Otkrist Gupta, Nikhil Naik, and Ramesh Raskar.
\newblock Designing neural network architectures using reinforcement learning.
\newblock In \emph{International Conference on Learning Representations}, 2017.

\bibitem[Bello et~al.(2017)Bello, Zoph, Vasudevan, and Le]{bello2017neural}
Irwan Bello, Barret Zoph, Vijay Vasudevan, and Quoc~V Le.
\newblock Neural optimizer search with reinforcement learning.
\newblock In \emph{International Conference on Machine Learning}, 2017.

\bibitem[Bender et~al.(2018{\natexlab{a}})Bender, Kindermans, Zoph, Vasudevan,
  and Le]{pmlr-v80-bender18a}
Gabriel Bender, Pieter-Jan Kindermans, Barret Zoph, Vijay Vasudevan, and Quoc
  Le.
\newblock Understanding and simplifying one-shot architecture search.
\newblock In Jennifer Dy and Andreas Krause, editors, \emph{Proceedings of the
  35th International Conference on Machine Learning}, volume~80 of
  \emph{Proceedings of Machine Learning Research}, pages 549--558,
  Stockholmsmässan, Stockholm Sweden, 10--15 Jul 2018{\natexlab{a}}. PMLR.
\newblock URL \url{http://proceedings.mlr.press/v80/bender18a.html}.

\bibitem[Bender et~al.(2018{\natexlab{b}})Bender, Kindermans, Zoph, Vasudevan,
  and Le]{bender2018demystifying}
Gabriel Bender, Pieter-jan Kindermans, Barret Zoph, Vijay Vasudevan, and Quoc~V
  Le.
\newblock Demystifying one-shot architecture search.
\newblock In \emph{International Conference on Machine Learning},
  2018{\natexlab{b}}.

\bibitem[Bengio et~al.(1994)Bengio, Bengio, and Cloutier]{bengio1994use}
Samy Bengio, Yoshua Bengio, and Jocelyn Cloutier.
\newblock Use of genetic programming for the search of a new learning rule for
  neural networks.
\newblock In \emph{Evolutionary Computation, 1994. IEEE World Congress on
  Computational Intelligence., Proceedings of the First IEEE Conference on},
  pages 324--327. IEEE, 1994.

\bibitem[Brock et~al.(2017)Brock, Lim, Ritchie, and Weston]{brock2017smash}
Andrew Brock, Theodore Lim, James~M Ritchie, and Nick Weston.
\newblock {SMASH}: one-shot model architecture search through hypernetworks.
\newblock In \emph{International Conference on Learning Representations}, 2017.

\bibitem[Chollet et~al.(2015)]{chollet2015keras}
Fran{\c{c}}ois Chollet et~al.
\newblock Keras, 2015.

\bibitem[Lee et~al.(2015)Lee, Zhang, Fischer, and Bengio]{lee2015difference}
Dong-Hyun Lee, Saizheng Zhang, Asja Fischer, and Yoshua Bengio.
\newblock Difference target propagation.
\newblock In \emph{Joint european conference on machine learning and knowledge
  discovery in databases}, pages 498--515. Springer, 2015.

\bibitem[Li et~al.(2017)Li, Jamieson, DeSalvo, Rostamizadeh, and
  Talwalkar]{li2017hyperband}
Lisha Li, Kevin Jamieson, Giulia DeSalvo, Afshin Rostamizadeh, and Ameet
  Talwalkar.
\newblock Hyperband: A novel bandit-based approach to hyperparameter
  optimization.
\newblock \emph{International Conference on Learning Representations}, 2017.

\bibitem[Liao et~al.(2016)Liao, Leibo, and Poggio]{liao2016important}
Qianli Liao, Joel~Z Leibo, and Tomaso~A Poggio.
\newblock How important is weight symmetry in backpropagation?
\newblock In \emph{AAAI}, 2016.

\bibitem[Lillicrap et~al.(2014)Lillicrap, Cownden, Tweed, and
  Akerman]{lillicrap2014random}
Timothy~P Lillicrap, Daniel Cownden, Douglas~B Tweed, and Colin~J Akerman.
\newblock Random feedback weights support learning in deep neural networks.
\newblock \emph{arXiv preprint arXiv:1411.0247}, 2014.

\bibitem[Linnainmaa(1970)]{lin70}
Seppo Linnainmaa.
\newblock The representation of the cumulative rounding error of an algorithm
  as a taylor expansion of the local rounding errors.
\newblock Master's thesis, 1970.

\bibitem[Loshchilov and Hutter(2016)]{loshchilov2016sgdr}
Ilya Loshchilov and Frank Hutter.
\newblock Sgdr: stochastic gradient descent with restarts.
\newblock \emph{arXiv preprint arXiv:1608.03983}, 2016.

\bibitem[N{\o}kland(2016)]{nokland2016direct}
Arild N{\o}kland.
\newblock Direct feedback alignment provides learning in deep neural networks.
\newblock In \emph{Advances in Neural Information Processing Systems}, pages
  1037--1045, 2016.

\bibitem[Ramachandran et~al.(2018)Ramachandran, Zoph, and
  Le]{ramachandran2018searching}
Prajit Ramachandran, Barret Zoph, and Quoc~V Le.
\newblock Searching for activation functions.
\newblock In \emph{International Conference on Learning Representations
  (Workshop)}, 2018.

\bibitem[Real et~al.(2017)Real, Moore, Selle, Saxena, Suematsu, Tan, Le, and
  Kurakin]{real2017large}
Esteban Real, Sherry Moore, Andrew Selle, Saurabh Saxena, Yutaka~Leon Suematsu,
  Jie Tan, Quoc Le, and Alex Kurakin.
\newblock Large-scale evolution of image classifiers.
\newblock In \emph{International Conference on Machine Learning}, 2017.

\bibitem[Real et~al.(2018)Real, Aggarwal, Huang, and Le]{real2018regularized}
Esteban Real, Alok Aggarwal, Yanping Huang, and Quoc~V Le.
\newblock Regularized evolution for image classifier architecture search.
\newblock \emph{arXiv preprint arXiv:1802.01548}, 2018.

\bibitem[Rumelhart et~al.(1986)Rumelhart, Hinton, and
  Williams]{rumelhart1986learning}
David~E Rumelhart, Geoffrey~E Hinton, and Ronald~J Williams.
\newblock Learning representations by back-propagating errors.
\newblock \emph{nature}, 323\penalty0 (6088):\penalty0 533, 1986.

\bibitem[Werbos(1974)]{wer74}
Paul Werbos.
\newblock \emph{Beyond Regression: New Tools for Prediction and Analysis in the
  Behavioral Sciences}.
\newblock PhD thesis, 1974.

\bibitem[Zagoruyko and Komodakis(2016)]{zagoruyko2016wide}
Sergey Zagoruyko and Nikos Komodakis.
\newblock Wide residual networks.
\newblock \emph{arXiv preprint arXiv:1605.07146}, 2016.

\bibitem[Zoph and Le(2017)]{zoph2016neural}
Barret Zoph and Quoc~V Le.
\newblock Neural architecture search with reinforcement learning.
\newblock In \emph{International Conference on Learning Representations}, 2017.

\bibitem[Zoph et~al.(2018)Zoph, Vasudevan, Shlens, and Le]{zoph2017learning}
Barret Zoph, Vijay Vasudevan, Jonathon Shlens, and Quoc~V Le.
\newblock Learning transferable architectures for scalable image recognition.
\newblock In \emph{Proceedings of the IEEE Conference on Computer Vision and
  Pattern Recognition}, 2018.

\end{thebibliography}

\newpage
\appendix






\section{Search space}
\label{appsearchspace}
The operands and operations to populate our search space are the following ($i$ indexes the current layer):

\begin{itemize}
    \item \textbf{Operands:}
    \begin{itemize}
        \item $W_i$, $\sgn(W_i)$ (weight matrix of the current layer and sign of it),
        \item $R_i$ , $S_i$ (Gaussian and Bernoulli random matrices, same shape as $W_i$),
        \item $R_{L i}$ (Gaussian random matrix mapping from $b^p_L$ to $b^p_i$),
        \item $h^p_i$, $h_i$, $h^p_{i+1}$ (hidden activations of the forward propagation),
        \item $b^p_L$, $b^p_{i+1}$ (backward propagated values),
        \item $b^p_{i+1} \frac{\partial h^p_{i+1}}{\partial  h_{i}}$, $b^p_{i+1} \frac{\partial h^p_{i+1}}{\partial  h^p_{i}}$ (backward propagated values according to gradient backward propagation),
        \item $b^p_{i+1} \cdot R_i$, $(b^p_{i+1} \cdot R_i) \odot \frac{\partial h_{i}}{\partial  h^p_{i}}$ (backward propagated values according to feedback alignment),
        \item $b^p_{L} \cdot R_{L i}$, $(b^p_{L} \cdot R_{L i}) \odot \frac{\partial h_{i}}{\partial  h^p_{i}}$ (backward propagated values according to direct feedback alignment).
    \end{itemize}
    \item \textbf{Unary functions $u(x)$:}
    \begin{itemize}
        \item $x$ (identity), $x^t$ (transpose), $1/x$, $\lvert x \rvert$, $-x$ (negation), $\mathds{1}_{x > 0}$ ($1$ if $x$ greater than $0$), $x^+$ (ReLU), $\sgn(x)$ (sign),
    $\sqrt{\lvert x \rvert}$, $\sgn(x) \sqrt{\lvert x \rvert}$, $x^2$, $\sgn(x) x^2$, $x^3$, $a x$, $x + b$,
        \item $x + \gnoise(g)$, $x \odot (1+\gnoise(g))$ (add or multiply with Gaussian noise of scale $g \in (0.01, 0.1, 0.5, 1.0)$),
        \item $\drop_d(x)$ (dropout with drop probability $d \in (0.01, 0.1, 0.3)$), $\clip_c(x)$ (clip values in range $[-c, c]$ and $c \in (0.01, 0.1, 0.5, 1.0$), 
        \item $\enorm{x}_0$, $\enorm{x}_1$, $\enorm{x}_2$, $\enorm{x}_{-inf}$, $\enorm{x}_{inf}$ (normalizing term by vector norm on flattened matrix),
        \item $\colnorm{x}_0$, $\colnorm{x}_1$, $\colnorm{x}_2$, $\colnorm{x}_{-inf}$, $\colnorm{x}_{inf}$,
    $\rownorm{x}_0$, $\rownorm{x}_1$, $\rownorm{x}_2$, $\rownorm{x}_{-inf}$, $\rownorm{x}_{inf}$,(normalizing term by vector norm along columns or rows of matrix),
        \item $\norm{x}_{fro}$, $\norm{x}_1$, $\norm{x}_{-inf}$, $\norm{x}_{inf}$ (normalizing term by matrix norm),
        \item $(x-\hat{m}(x)) / \sqrt{\hat{s}(x^2)}$ (normalizing with running averages with factor $r = 0.9$).
    \end{itemize}
    \item \textbf{Binary functions $f(x, y)$:}
    \begin{itemize}
        \item $x + y$, $x - y$, $x \odot y$, $x / y$ (element-wise addition, subtraction, multiplication, division),
        \item $x \cdot y$ (matrix multiplication),
        \item $x$ (keep left),
        \item $\min(x, y)$, $\max(x, y)$ (minimum and maximum of $x$ and $y$).
    \end{itemize}
\end{itemize}

Additionally, we add for each operand running averages for the mean and standard deviation as well as a normalize version of it, i.e., subtract the mean and divide by the standard deviation.

So far we described our setup for dense layers.
Many state-of-the-art neural networks are additionally powered by convolutional layers, therefore we chose to use convolutional neural networks.
Conceptually, dense and convolutional layers are very similar, e.g., convolutional layers can be mimicked by dense layers by extracting the the relevant image patches and performing a matrix dot product.
For performance reasons we do to use this technique, but rather map the matrix multiplication operations to corresponding convolutional operations.
In this case we keep the (native) $4$-dimensional tensors used in convoluational layers and, when required, reshape them to matrices by joining all axes but the sample axis, i.e., join width, height, and filter axes.

\section{Experimental details}
\label{appexperimentaldetails}


The Wide ResNets used in the experiments are not just composed of dense and convolutional layers, but have operations like average pooling.
For parts of the backward computational graph that are not covered by the search space, we use the standard gradient equations to propagate backwards.


Throughout the paper we use the CIFAR-10 dataset and use the preprocessing  described in~\citet{loshchilov2016sgdr}.
The hyperparameter setup is also based on that in~\cite{loshchilov2016sgdr}, and we only modify the learning rate and the optimizer.
For experiments with more than $50$ epochs, we use cosine decay with warm starts, where warm up phase lasts for $10\%$ of the training steps and the cosine decay reduces the learning rate over the remaining time.

The evolution process is parameterized as follows.
With probability $p = 0.7$ we choose an equation out of the $N = 1000$ most competitive equations.
For all searches we always modify one operand or one operation.
We use Keras~\citep{chollet2015keras} and Tensorflow~\citep{abadi2016tensorflow} to implement our setup.
We typically use 500 workers which run on CPUs for search experiments, which lead to the search converging within 2-3 days.

We experiment with different search setups, i.e., found it useful to use a different subset of the available operands and operations,
to select different learning rates as well as optimizers.
We use either WRNs with depth 16 and width 2 or depth 10 and width 1 and use four different learning rates.
Additionally, during searches we use early stopping on the validation set to reduce the overhead of bad performing equations. Either SGD or SGD with momentum (with the momentum coefficient set to $0.9$) is used as the optimizer.
The controller tries to maximize the accuracy on the validation set. The top $100$ equations are trained on the entire training set, and the test set accuracy is reported in Table~\ref{table}.



\end{document}